\documentclass[conference]{IEEEtran}
\usepackage{times}

\usepackage{multicol}
\usepackage[bookmarks=true]{hyperref}
\usepackage{amsmath} 
\usepackage{amssymb}  
\usepackage{algpseudocode}
\usepackage{pifont}
\usepackage{xcolor}
\usepackage{soul}
\usepackage{lipsum}
\usepackage{booktabs}
\usepackage{adjustbox}
\usepackage{mathtools}
\usepackage{float}
\usepackage{esvect}
\usepackage{nicefrac, xfrac}

\newcommand{\bbezier}{{B\'ezier }}
\newcommand{\gaction}{{u}}

\newcommand{\state}{{\mathbf{x}}}

\newcommand{\particles}{{\boldsymbol\theta}}
\newcommand{\particle}{{\theta}}
\newcommand{\obs}{{\mathbf{z}}}

\newcommand{\stiffness}{{\mathbf{k}}}
\newcommand{\p}{{\mathbf{p}}}

\newcommand{\curve}{{S}}

\newcommand{\argmin}{\mathop{\mathrm{argmin}}}

\definecolor{ForestGreen}{rgb}{0.133, 0.545, 0.133}
\usepackage[ruled,vlined,linesnumbered]{algorithm2e}

\SetCommentSty{mycommfont}

\begin{document}

\title{\textbf{ProCut}: \textbf{Pro}babilistic \textbf{Cut}ting Topology for Autonomous Electrosurgical Tissue Dissection}

\author{Xiao Liang, Fei Liu, Florian Richter, Genie Rubaiyat, Michael Yip}

\maketitle


\begin{abstract}
Accurately modeling and tracking the deformation of soft tissue is critical for a wide range of interventional and surgical procedures. However, current methods struggle in scenarios involving topological changes—such as cutting and dissection—due to the inherent non-linearity and discontinuity introduced by explicit changes in connectivity. In this work, we present a novel, fully differentiable framework that enables robust estimation and modeling of topological changes during deformable tracking. Our method introduces a continuous, sigmoid-based formulation to smooth the otherwise discrete event of tissue cutting, making it amenable to gradient-based optimization within a differentiable Position-Based Dynamics (PBD) simulation. To account for uncertainty and improve robustness in the presence of noisy visual data, we incorporate Stein Variational Gradient Descent (SVGD) for particle-based probabilistic inference, generating multiple hypotheses for topological state estimation. Building on this foundation, we develop an autonomous dissection algorithm for thin-shell tissues that leverages topological updates to guide closed-loop cutting trajectory control. We evaluate our approach in both simulated and real-world electrosurgical environments, demonstrating significant improvements in topological estimation accuracy and dissection precision over existing methods. Our results highlight the potential of this framework to advance automation in soft-tissue surgical procedures by enabling reliable perception and control in the presence of complex structural changes.
\end{abstract}

\section{INTRODUCTION}
Modeling and tracking of deformable objects, such as soft tissue and human organs, play a crucial role in various interventional and surgical procedures \cite{schmidt2024tracking}. The process includes capturing soft bodies deformation \cite{gong2024self} and enabling understanding and prediction of physical behaviors of the deformable objects over time \cite{liang2024real}. 
In interventional procedures with CT and MRI data, deformable tracking methods elevate the reliability of procedures such as radiation therapy \cite{10.1371/journal.pone.0271064} and image-guided biopsy \cite{nakamura2021computed}.
In image-guided surgery, this technology assists surgeons by providing additional visual guidance \cite{bernhardt2017status}, reducing their cognitive load. Recent efforts in surgical autonomy such as tissue manipulation \cite{hu2023occlusion}, blood suction \cite{Florian_2021_RAL}, and suturing \cite{kam2021confidence}, have also relied on deformable tracking algorithms applied to endoscopic image data, serving as crucial cues for the autonomous agents.
Despite being broadly used, a significant limitation remains: existing deformable modeling and tracking methods struggle to handle tissue connectivity changes that is ubiquitous in surgery.
This limitation poses a challenge for further advancement in bringing autonomy to surgery including accurate tissue visualization after dissections or even autonomy of tissue dissection itself.

Most prior methods in deformable tracking and modeling rely on visual inputs,
whether optimization-based \cite{10160746, saha2023based} or learning-based \cite{liang2023orrn, gong2024self}, commonly incorporate spatial smoothness objectives to encourage physically plausible deformation estimation.
However, these computer vision techniques do not explicitly consider topological changes leading to over-regularizing and inaccuracies in regions with topological changes.
Other works integrates visual observations with physics-based modeling, including physics-based deformable registration \cite{werner2009patient} and real-to-sim state estimation \cite{sundaresan2022diffcloud}.
These approaches are also limited by their underlying physics models, which are not designed to accommodate topological changes.
Explicit estimation of topological changes has been explored through direct deletion of mesh edges based on a heuristic-strategy \cite{paulus2017handling, wang2025feedback} and learning approaches \cite{franccois2021image}.
However, these approaches are limited in robustness due to their sensitivity to observation noise, reliance on handcrafted rules or training data, and reduced generalization ability to unseen topological variations.
\begin{figure}[t]
    \centering
    \includegraphics[width=0.98\linewidth]{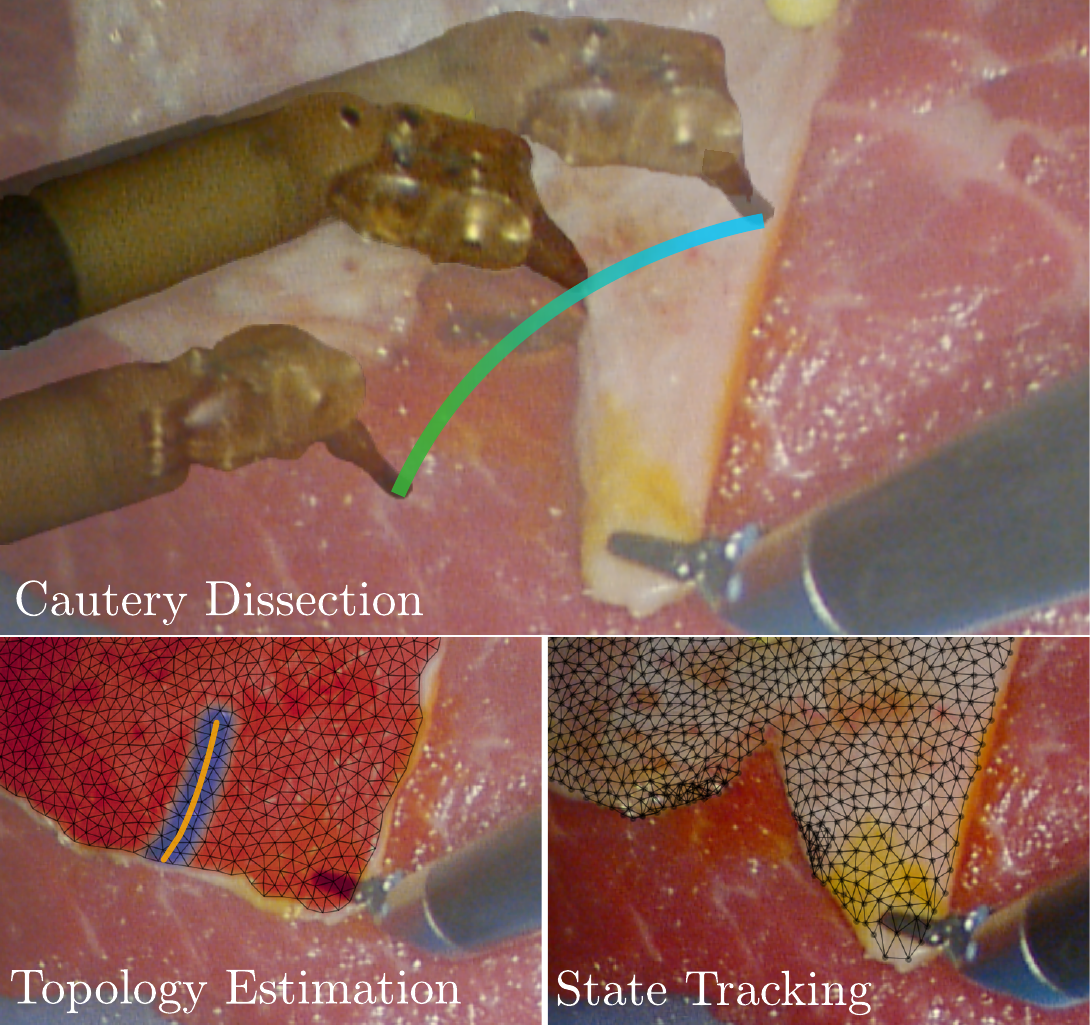}
    \vspace{-0.5em}
    \caption{Overview of our autonomous tissue dissection system. (Top) The robot performs electrosurgical dissection. (Bottom left) Our method estimates the cutting topology using probabilistic, gradient-based inference on a canonical reference frame. (Bottom right) The estimated topology enables accurate tracking of tissue deformation, providing critical feedback for precise dissection control.}
    \vspace{-1.5em}
    \label{fig:autonomous_cutting_sim}
\end{figure}

To overcome these limitations, we propose \textbf{ProCut}, a differentiable formulation for tracking, modeling, and applying topological changes during tissue cutting tasks.
Tissue cutting introduces a highly non-linear transformation, as it involves explicit removal of mesh connectivity.
Our approach makes this non-linearity amenable to gradient-based optimization by smoothing the connectivity change using a sigmoid function.
formulation is integrated into a differentiable physics simulator, Position-Based Dynamics (PBD), for deformable tracking and modeling \cite{liang2024real}.
Additionally, we incorporate the estimation of topological changes into a particle-based probabilistic estimation algorithm, Stein Variational Gradient Descent (SVGD), enabling multiple hypotheses to initialize the optimization process and providing uncertainty of the estimated topological change.
Leveraging this fully differentiable framework,  an autonomous dissection algorithm is devised that is tailored to thin-shell tissue, using topological updates as cues for closed-loop dissection trajectory control. In summary, the main contributions of this work are:
\begin{itemize}
    \item \textbf{A differentiable cutting topology model} that represents tissue
    connectivity through a smooth, gradient-compatible formulation, enabling
    end-to-end optimization of dissection trajectories within a differentiable simulator.
    \item \textbf{A probabilistic topology estimator} based on SVGD that maintains a distribution over cutting curves
    from visual observations, providing calibrated uncertainty estimates to
    guide closed-loop dissection under deformation and observation noise.
    \item \textbf{An autonomous electrosurgical dissection pipeline}
    integrating the above components into a real-time perception-planning-control
    loop, validated in both simulation and on ex-vivo tissue, and demonstrating
    recovery from topology estimation errors without replanning from scratch.
\end{itemize}

\section{RELATED WORK}
\textbf{Deformable Cutting Simulation.}
The simulation of soft body cutting has been extensively studied using methods such as rope-grid \cite{haiderbhai2024simulating}, PBD \cite{yu2024real, berndt2017efficient}, Finite Element Methods (FEM) \cite{wu2005improved} \cite{scheikl2023lapgym, schegg2023sofagym}, and the Material Point Method (MPM) \cite{xu2023roboninja}. These works provide valuable testbeds, but they
are designed as forward simulators, but lacks the ability to incorporate real-world feedback.
A second line of work has introduced differentiability into cutting
simulation to enable gradient-based optimization.
SliceIt!~\cite{beltran2024sliceit} proposes a differentiable simulator for
training cutting agents, but similarly does not address topology estimation
from real observations.
DiSECt~\cite{heiden2021disect} is the most closely related effort: it
formulates cutting as the gradual weakening of virtual springs connecting
nodes across a \emph{predefined cutting plane}, which restricts the geometry
of admissible cuts. In contrast, our method introduces a differentiable
\emph{cutting-curve} representation that directly parameterizes the cut path
on the tissue surface, allowing the topology itself to be an optimization
variable inferred from visual feedback, which is essential for closed-loop dissection.
\cite{paulus2017handling} combined visual feature tracking with an FEM-based tissue simulation to update the model topology.
Nevertheless, this approach relies on hand-crafted heuristics to detect topological changes, making it susceptible to noise in observational data. 

\textbf{Autonomous Cutting and Dissection.} 
\cite{9215039} proposed a vision-based closed-loop controller to ensure accurate knife cutting depth along a predefined trajectory. \cite{nguyen2019manipulating, shahkoo2023autonomous} developed reinforcement learning algorithms to simultaneously perform tensioning and cutting on surgical gauze pads. 
Electrosurgical dissection is more often used in tissue dissection in robotic surgeries. 
For instance, \cite{saeidi2019supervised} experimented with a near-infrared marking method for electrosurgery on porcine tongue samples.
\cite{ge2021supervised} introduced a suction gripper to retract tissue, facilitating electrocautery procedures. 
Similarly, \cite{liang2024medic} focused on dissection assistance that helps reveal the surgical site during electrocautery procedures.
\cite{oh2023framework} proposed a framework that leverages semantic segmentation and key-point detection for electrocautery along tissue boundaries. 
Furthermore, \cite{karimi2024reward} designed a reinforcement learning-based approach to control a cautery tool to remove tissue attachments.
While these methods demonstrate progress in automating Electrosurgical dissection tasks, they are largely constrained to controlled environments and do not account for real-world challenges such as noise, disturbances, and changing tissue topology.

\section{Preliminaries}

ProCut builds off previous work in differentiable simulation for tissue modeling and tracking \cite{liang2024real}.
Let $M_t: (\state_t, \mathbf{F})$ be a mesh of a thin-shell soft tissue where $\state_t = \{x^i_t\}^n_{i=1}$ is a set of vertices in $\mathbb{R}^3$, and set $\mathbf{F}=\{F^j\}_{j=1}^m$ contains all triangle faces of the mesh. Each triangle face contains three vertices $F^j = (i1, i2, i3)$.
We denote $\state_t = f'(\state_{t-1}, u_t, \stiffness, \obs_t)$ as our forward simulation step where $\gaction_t$ is the applied control to the tissue, $\stiffness$ defines the stiffness of the thin-shell soft tissue, and $\obs_t$ is the surface point cloud observation of the real tissue. 

\begin{algorithm}[!t]
    \caption{Autonomous Tissue Dissection}
    \label{alg:feedback_control}
    \SetKwInOut{Input}{Input}
    \SetKwInOut{Output}{Output}

    \Input{\text{Planned cutting path} $\mathcal{\hat T}=\{\hat\tau_i\}_{i=1}^{T}$, \text{Initial and tensioning grasping point} $\gaction_0, \gaction_\text{ten}, \text{Cautery tooltip pose }\mathbf{T}$}

        Grasp tissue at $\gaction_0$ and move to tensioning position $\gaction_\text{ten}$

        $\mathbf{S} \gets \{\}$



        $\hat \tau_{s}, \hat \tau_{e} \gets \hat \tau_1, \hat\tau_2$

        \For {$i$ \text{in} [1,...,T]} {

            \tcp{Execute a dissection action}
            
            $\{\hat w_j\}_{j=0}^{J} \gets \text{linearInterpolate}(\hat\tau_s, 
            \hat\tau_e)$

            $\{w_j\} \gets g^{-1}_{\state_0}(\{\hat w_j\}, \state_{i-1})$
            
            \If {$i=1$} {
                $\text{Move tooltip to } w_0$
            }
        
            \For {$j$ in $[1,...,J]$} {

                $\delta w \gets  w_j - w_{j-1}$

                $\mathbf{R} \gets \mathbf{R} \textbf{ s.t. } \frac{\mathbf{R}.\vv{x}}{|\mathbf{R}.\vv{x}|} = \frac{\delta w}{|\delta w|} \text{ and } \mathbf{R}.\vv{z} \perp \text{localMeshSurface}(w_j)  $

                $\text{Move relative }\delta w$ \text{with zigzag pattern, }
                \text{keep orientation}$\mathbf{R}$
            }

            \tcp{Estimate cutting outcome}
            
            $\bar S_i \gets \text{initCuttingCurve}(\hat \tau_s, \hat \tau_e)$
    

            $S_i, \state_i \gets \text{cutEstimation}(\bar S_i, \mathbf{S}, \state_{i-1}, \gaction_\text{ten})$

            $\mathbf{S} \gets \mathbf{S} + \{S_i\}$

            \tcp{Update cutting plan}

            \If {$i < T$} {

                $\text{Cutting end point } c_e \gets q_e \text{ of } S_i$
                $\hat \tau_s, \hat \tau_e \gets c_e, \tau_{i+1}$
            
            }
            
        }

\end{algorithm}

We use a quasi-static Extended Position-based Dynamics (XPBD) simulation to model thin-shell soft tissues.
The simulator is based on geometric constraints which directly update the position of the thin-shell soft tissue through constraint satisfaction.
The two geometric constraints used in this work are distance to simulate springs and an observational real-to-sim constraints.
The distance constraints is defined as
\begin{equation}
    \begin{split}
        C_d(x^i, x^j) = \|x^i-x^j\| - \|x^i_0-x^j_0\|,\ (i, j) \in \mathbf{L}(M).
    \end{split}
\end{equation}
$\mathbf{L}(M)$ is the connectivity graph of the mesh. It contains all pairs of vertices that are connected by an edge, as well as all pairs of non-neighboring vertices of two adjacent triangle faces for simulating pseudo bending constraints \cite{zhang2023achieving}. Every distance constraint $C_d(x^i, x^j)$ is associated with a stiffness value $k^{i,j} \in \stiffness$. Note that we adopts a variable stiffness formulation which is useful to model the non-linear dissection effects later.
The real-to-sim constraint enforces similarity between mesh vertices and the current observed point cloud. 
It is defined by the chamfer distance between mesh vertices and the tissue point cloud
\begin{equation}
    \begin{split}
        C_r(\state, \obs) &= d_\text{chamfer}(\state, \obs)
        \\ &=\sum_{x\in \state} \min_{z\in \obs}\|x - z\|^2_2  + \sum_{z\in \obs} \min_{x\in \state}\|z -  x\|^2_2.\\
    \end{split}\label{eqn:chamfer}
\end{equation}
This constraint encourages the similarity between the simulation and the visual observation.

In our simulation, we also incorporate a robotic gripper which grasps and directly controls a portion of the thin-shell soft tissue.
The gripper's position, $\gaction_t$, is assumed to stay grasping to the same location the entire time series.
Therefore, we search for the k-th nearest vertices in $\state_0$ closest to $\gaction_0$ and apply a displacement is force equal to $\gaction_t - \gaction_0$ to those vertices everytime simulation timestep.
Putting it all together, our forward simulation step
is solved for with the following optimization problem:
\begin{equation}\label{eqn:energy_minimization}
\begin{split}
\state_{t} &= \argmin_{\state}\sum_{i,j\in \mathbf{L}(M)} \frac{1}{2} k^{i, j} C_d(x^i, x^j)^2  + C_r(\state, \obs_t)\\
&\textbf{s.t.}\ x^l - x_0^l = \gaction_t - \gaction_0,\ \forall x_0^l \in N^K_{\state_0}(\gaction_0)
\end{split}
\end{equation}
where $N^K_{*}$ finds the $k$ nearest vertices.
Our simulator solves the optimization problem iteratively using Gauss-Seidal method.
For convenience later, we introduce a variant of our simulation step $f(\state_{t-1}, \gaction_t, \stiffness)$ that solves \autoref{eqn:energy_minimization} but without $C_r$ terms.
We directly differentiate through the optimization iterations to solve for any partial derivatives, $\frac{\partial f}{\partial \textbf{x}_{t-1}}, \frac{\partial f}{\partial u}_{t}, \frac{\partial f}{\partial \textbf{k}}$, providing a differentiable simulator.

\section{Methods}\label{sec:methods}
\begin{figure}[t]
    \centering
    \includegraphics[width=1\linewidth]{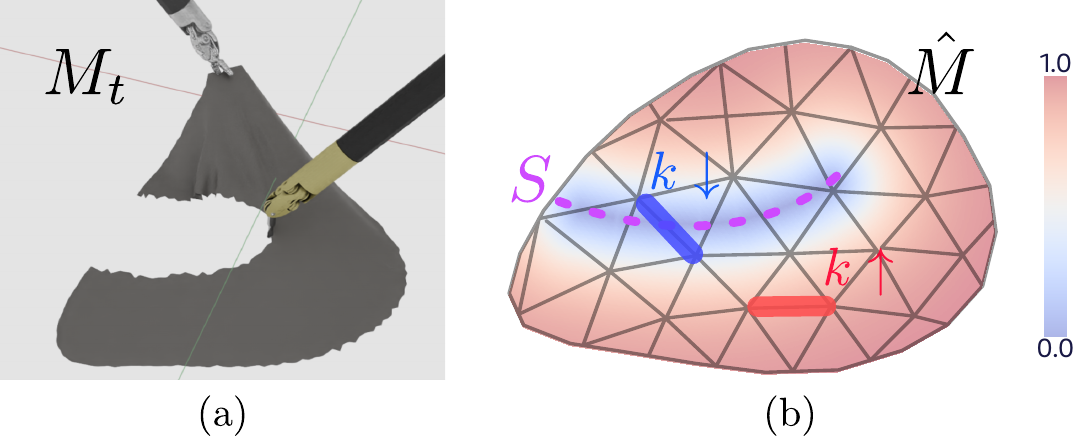}
    \vspace{-2em}
    \caption{Key representations that enable differentiable optimization of the cutting topology. (a). A mesh $M_t$ representing tissue that is manipulated and cut by a robot. (b). In the tissue's canonical mesh representation $\hat M$, a cutting curve $S$ affects the constraint stiffness of its surrounding mesh edges. Edges that are closer to the curve will have smaller stiffness to achieve similar effect as if the edges are cut.}
    \vspace{-1.5em}
    \label{fig:cut_parametrization}
\end{figure}

We parameterize a cut as $S$, and integrate it into our simulation to satisfying geometric constraints that makes the mesh behaves like real soft tissues.
Furthermore, we ensure the cut is end-to-end differentiable, i.e. $\frac{\partial f}{\partial S}$ can be computed, hence integrating topological changes in a differentiable manner.
We leverage the differentiability of the topological changes into our real-to-sim framework for tissue which ultimately leads to an algorithm that solves two optimization problems simultaneously from observed point cloud: one that estimates the mesh state, $\textbf{x}_t$ and another that estimates the posterior probability of the cut.
Alg.~\ref{alg:cut_estimation} gives an overview of the estimation algorithm. 
We present an autonomous algorithm summarized in Alg.~\ref{alg:feedback_control} that utilizes the estimation results for doing feedback control.

\subsection{Canonical Representation and Cutting Mechanism} \label{sec:representation}
Our framework relies on a simple intuition that tissue dissection outcome can be modeled by reducing the regional distance constraints stiffness at the dissected area.
For example, an edge has a zero stiffness value if it is ``dissected" so its distance constraint no longer needs to be satisfied in Eqn.~\ref{eqn:energy_minimization}. We parameterize outcomes of a cut as a curve that lies on the manifold of the canonical mesh $\hat M: (\mathbf{\hat x}, \mathbf{F})$, as illustrated in Fig~\ref{fig:cut_parametrization}.
The cutting curve affects the stiffness parameters such that edges that are close to the curve have smaller stiffness, and have larger stiffness values otherwise.

\textbf{Canonical representation:}
We choose a mesh's planar parametrization as its canocial representation. Specifically, we use the Least-squared Conformal Mapping (LSCM) planar parameterization technique to construct a canonical representations. At time $t=0$, let $g: \mathbb{R}^3 \rightarrow \mathbb{R}^2$ be the LSCM parameterization that maps 3D mesh vertices $\mathbf{x}_0$ to 2D planar coordinates $\mathbf{\hat x} = g(\mathbf{x}_0)$. 
Given points $\mathbf{\hat p} \in \hat M$ on the canonical 2D domain, we define an inverse map $g^{-1}_{\state_0}(\mathbf{\hat p}, \mathbf{x}_t)$ that retrieves the corresponding point on the deformed mesh at time $t$, based on its barycentric coordinates within the triangle it lies in. The inverse mapping is defined as: 
\begin{equation}\label{eqn:barycentric_coordinate}
    \begin{split}
        g^{-1}_{\state_0}(\mathbf{\hat p}, \mathbf{x}_t) = \begin{bmatrix}
            x_0^{i1} & x_0^{i2} & x_0^{i3}
        \end{bmatrix} \cdot \boldsymbol{\lambda}(\state_0, \mathbf{\hat p})
    \end{split}
\end{equation}
where $(i1, i2, i3)$ are the vertex indices of the traingle $F^j$ in which $\mathbf{\hat p}$ lies, and $\boldsymbol\lambda \in (0, 1)^3$ are computed as
\begin{equation}
    \begin{split}
        \boldsymbol{\lambda}(\state_0, \mathbf{\hat p}) = \begin{bmatrix}
            g(x_0^{i1}) & g(x_0^{i2})&g(x_0^{i3})\\
            1&1&1\\
        \end{bmatrix}^{-1}\begin{bmatrix}
            \mathbf{\hat p} \\ 1
        \end{bmatrix}.
    \end{split}
\end{equation}
%

\textbf{Cutting mechanism:} 
We define a cutting curve $S \subseteq \hat M$ on the canonical mesh and model the spatially-varying stiffness as a function of distance to this curve. For each distance constraint pair $(i, j)$, the stiffness is defined as:
\begin{equation}\small
    k^{i,j}(\curve) = k \cdot \sigma \left( \frac{ \mathcal{D} \left( \frac{1}{2}(g(x^i_0 + x^j_0)), \curve\right)-r}{h} \right), k^{i,j}(\curve)\in \stiffness(\curve)
\end{equation}
where $D(\cdot, S)$ is a distance function to a curve $\curve \in\mathbf{S}$ which is a prart of the cut. $\sigma(\cdot)$ denotes a sigmoid function, and  $r, h$ are tunable hyper-parameters controlling the decay of stiffness.
This formulation supports multiple cutting curves on the same tissue by selecting the minimum stiffness from each curve.
Given a set $\mathbf{S}=\{\curve_l\}_{l=1}^T$, the final stiffness is
\begin{equation}
    \begin{split}
        k^{i, j}(\mathbf{S}) = \min_l k^{i, j}(S_l). 
    \end{split}
\end{equation}
Our formulation based on reducing the stiffness associated with a cutting region to is fully differentiable, i.e. $\frac{\partial k^{i,j}}{\partial \textbf{S}}$ exists.
Therefore, the full gradient for the differentiable simulation is be computed via chain rule, $\frac{\partial f}{\partial \textbf{S}} = \frac{\partial f}{\partial k^{i,j}} \frac{\partial k^{i,j}}{\partial \textbf{S}}$ where $\frac{\partial f}{\partial k^{i,j}}$ is computed in the same manner as our prior work \cite{liang2023realtosim}.

\textbf{Curve Parameterization:} 
We explore both linear and quadratic \bbezier to parametrize the cutting curve. A linear \bbezier curve is represented with two control points $\curve^{\text{linear}}:\{q_s, q_e\}$, with path $\mathcal{B}^\text{linear}(t, \curve^{\text{linear}}) =q_s + t(q_e - q_s),  0 \le t \le 1$. The minimum distance from a point $x \in \mathbb{R}^2$, its minimal distance to curve to this line segment is
    \begin{equation}\label{eqn:line_segment_distance_function}
        \begin{split}
            A = x - q_e, B &= q_s - q_e, C= x-q_s,\ t = \frac{\sqrt{A \cdot B}}{\|B\|}\\
            D(x, \curve^{\text{linear}}) &=
            \begin{cases}
                \|A\|,& t < 0 \\
                \| x - \mathcal{B}^\text{linear}(t,\curve^{\text{linear}}) \|,& 0 \le t \le 1\\ 
                \|C\|,& t > 1\\
            \end{cases}
        \end{split}
    \end{equation}
    
For quadratic \bbezier curves defined by three control points $\curve^\text{quad}:\{q_s, q_m, q_e\}$, the path is:
\begin{equation}
    \begin{split}
        \mathcal{B}^\text{quad}(t, \curve^\text{quad}) = (1-t)^2 q_s + 2(1-t)&tq_m + t^2 q_e, \\ 
        &0 \le t \le 1\\
    \end{split}
\end{equation}
Though an analytical distance function exists, we found it numerically unstable during back-propagation in practice. Instead, we approximate its distance function by discretizing quadratic \bbezier curve into line segments and composing their distance functions together: 
\begin{equation}
    \begin{split}
        t^i & = \frac{i}{n}, i \in \{0,1,...,n-1\}\\
        \curve_{i}^{\text{linear}} &= \{\mathcal{B}^\text{quad}(t^i),\mathcal{B}^\text{quad}(t^{i+1})\},  \\
        D(x, \curve^\text{quad})& = \min_i D(x, \curve_{i}^{\text{linear}}),  \\
    \end{split}
\end{equation}
where $n$ is the number of line segments that we tune. This approximation not only stabilizes optimization but also enables use of higher-degree curves. In practice, we leverage both linear and quadratic representations by first bootstrapping the estimation with a linear curve and then refining it to a quadratic curve after a set number of iterations.

\begin{algorithm}[!t]
    \caption{Cutting Curves Estimation}
    \label{alg:cut_estimation}
    \SetKwInOut{Input}{Input}
    \SetKwInOut{Output}{Output}

    \Input{\text{Mesh state} $\state_{t-1}$, \text{cutting curve prior} $\bar \curve_t$, previous cutting curves $\mathbf{S}$, gripper position $\gaction$, \text{iteration} $I$, number of particles $N$.}
    
        $\particles \gets \{\text{UniformSample}(\bar \curve_t)\ N \text{ times}\}$

        $\state_{i=0} = \state_{t-1}$
        
        \For{i in $[1,...,I]$}{

            $\obs_t \gets \text{stereoReconstruct}()$
            
            \tcp{SVGD update step}
            
            $\mathcal{\bar X}_{i-1} \gets \{\text{repeat }\state_{i-1}\ N \text{ times}\}$
            
            $\boldsymbol\theta \gets \text{ParticleUpdate}(\boldsymbol\theta, \mathbf{S}, \mathcal{\bar X}_{i-1}, \gaction, \obs_t)$

            \tcp{Update mesh states}
            
            \For{j in $[1,...,N]$} {
                $\mathcal{\bar X}_i[j] \gets f'(\mathcal{\bar X}_{i-1}[j], \gaction, \stiffness(\{\theta^j\} + \mathbf{S}), \obs_t)$
            }
            
            $l \gets \argmin_{l} C_r(\mathcal{\bar X}_i[l], \obs_t)$
            
            $ \state_i \gets \mathcal{\bar X}_i[l]$
            
        }

        $\mathbf{S} \gets \mathbf{S} + \boldsymbol{\theta}[l]$

        $\textbf{return}\ \boldsymbol{\theta}[l], \mathcal{\bar X}_i[l]$
        
\end{algorithm}

\subsection{Cutting Curve Estimation} \label{sec:estimation}

We estimate the curve parameters, $\mathbf{S}$, from point cloud data in a probabilistic fashion.
The posterior $\p(\mathbf{S}|\obs_t)$ over cutting curve parameters and the surface tissue observation can be computed from Bayes' rule $\p(\mathbf{S}|\obs_t) \propto \p(\mathbf{S})\p(\obs_t|\mathbf{S})$ where the $\p(\mathbf{S})$ is the prior distribution over the cutting curves.
The likelihood term $\p(\obs_t|\mathbf{S})$ is computed the physics model that maps cutting parameters to expected observations.
Specifically, we model the observation likelihood with a forward simulation based on a known previous tissue state $\state_{t-1}$, leading to a predicted hidden state $\bar \state_t$. The likelihood becomes:
\begin{equation}
    \begin{split}
        \p(\obs_t | \mathbf{S}) &= \p(\obs_t|\bar \state_t)\\&= \p\Bigl(\obs_t|f\left(\state_{t-1}, \gaction_{t}, \stiffness(\mathbf{S})\right)\Bigr) 
    \end{split}
\end{equation}
The observation model represents the likelihood of a surface point cloud given a simulated mesh state. Inspired by \cite{5432191}, we consider mesh vertices $\bar \state_t$ as centroids of a Gaussian Mixture Model (GMM), and surface point cloud $\obs_t$ as data generated by the GMM. The density function is 
\begin{equation}\label{eqn:original_gmm}
\begin{split}
\p(z|\bar\state_t) &= \sum_{x \in \bar\state_t} \frac{1}{n}\mathcal{G}(z | x, \sigma),z \in \obs_t\\
\mathcal{G}(z | x, \sigma) &=  \frac{1}{\sigma \sqrt{2\pi}} e^{-\frac{\|z - x\|^2}{2\sigma^2}}.
\end{split}
\end{equation}
In the above equation, individual effect of GMM centroids become smaller when two points are further apart. Therefore, for computation efficiency, we further approximate it to only consider the K-nearest GMM centroids
\begin{equation}\label{eqn:modifieds_gmm}
\begin{split}
\p(z|\bar\state_t) &\approx \sum_{x \in N^K_{\bar \state_t}(z)} \frac{1}{K}\mathcal{G}(z | x, \sigma).\\
\end{split}
\end{equation}
Finally, the observation model is computed $\p(\obs_t|\bar \state_t)=\prod_{z\in\obs_t}\p(z|\bar\state_t)$.

We use a set of particles $\particle^i \in \particles$ to represent the posterior distribution of cutting curves due to their flexibility in capturing a wide range of distributions rather than a parametric model. 
To update the particles and solve for the distribution, we use Stein Variation Gradient Descent (SVGD). It has the following update rule 
\begin{equation}\label{eqn:svgd_update}
    \begin{split}
       \theta^i  \gets \theta^i + \epsilon \phi^*( \theta^i)&\\
        \phi^*(\theta) = \frac{1}{m}\sum_{j=1}^{m}h( \theta^j,  \theta)& \nabla_{ \theta^j}\log \p( \theta^j|\obs_t) \\
         &+ \nabla_{ \theta^j}h( \theta^j,  \theta).\\
    \end{split}
\end{equation}
$h(\theta,\theta')=\exp\bigl(-\frac{(\theta-\theta')^2}{2l^2}\bigr)$ represents a Radial Basis Function (RBF) kernel applied to the flattened curve parameters $\bar \theta$.

The mesh state $\state_t$ is updated iteratively during Bayesian optimization by simulating tissue states in parallel with $f'$ across particles and selecting the one that best satisfies the real-to-sim constraint in \autoref{eqn:chamfer}. To prevent exponential growth in particle count as new cutting curves are added to $\mathbf{S}$, we maintain a fixed number of particles by greedily selecting the best one using the same heuristic. Each new particle set inherits prior knowledge and focuses on representing the most recent cutting curve.

\subsection{Autonomous Tissue Dissection}\label{sec:feedbackcontrol}

We develop an autonomous dissection algorithm for thin-shell tissue, outlined in Alg.~\ref{alg:feedback_control}, which leverages cutting curve estimation to guide robot actions that closely follow the target trajectory while respecting tissue deformation

Given a goal trajectory $\mathcal{\hat T} = {\hat \tau_i}$ on the canonical mesh $\hat M$, each robot action dissects a segment between two waypoints.
After performing an action, a new cutting curve is estimated (line 13-15). Let $\hat \tau_s$ be a known starting point, or an estimated ending point of a previous cutting curve, the estimation algorithm uses a prior for initialization: 
\begin{equation}
    \begin{split}
        \bar S_{i} =
        \begin{cases}
            \{\hat \tau_s, \hat \tau_i\}, & \text{if}\ linear,\\
            \{\hat \tau_s, \frac{\hat \tau_s + \hat\tau_i}{2}, \hat\tau_i\},&\text{if}\ quad,
        \end{cases}
    \end{split}
\end{equation}
where \textit{linear} and \textit{quad} are chosen depending on the curve representation.
The estimated cutting curve facilitates more accurate tissue tracking and a controller updates future actions using the new curve’s endpoint to reduce trajectory deviation. (line 16-17).

We uses velocity control to execute each robot action (line 5-12). 
To compute velocities, a segment is further interpolated into smaller way-points $\{\hat w_j\}_{j=1}^W$. These are mapped to the deformed mesh using \autoref{eqn:barycentric_coordinate}, and end-effector velocities are computed as differences between consecutive waypoints
Then, the robot's end-effector velocities in Cartesian's space is calculated as differences between every two smaller way-points. 
To ensure robust dissection, we add motion along the tissue’s normal direction, producing a zig-zag pattern, and constrain the tool tip to point against the surface normal. Throughout the whole cutting process, the grasping point remains fixed at a predefined tensioning position $u_{\text{ten}}$.

\begin{figure}[t]
    \centering
    \includegraphics[width=1\linewidth]{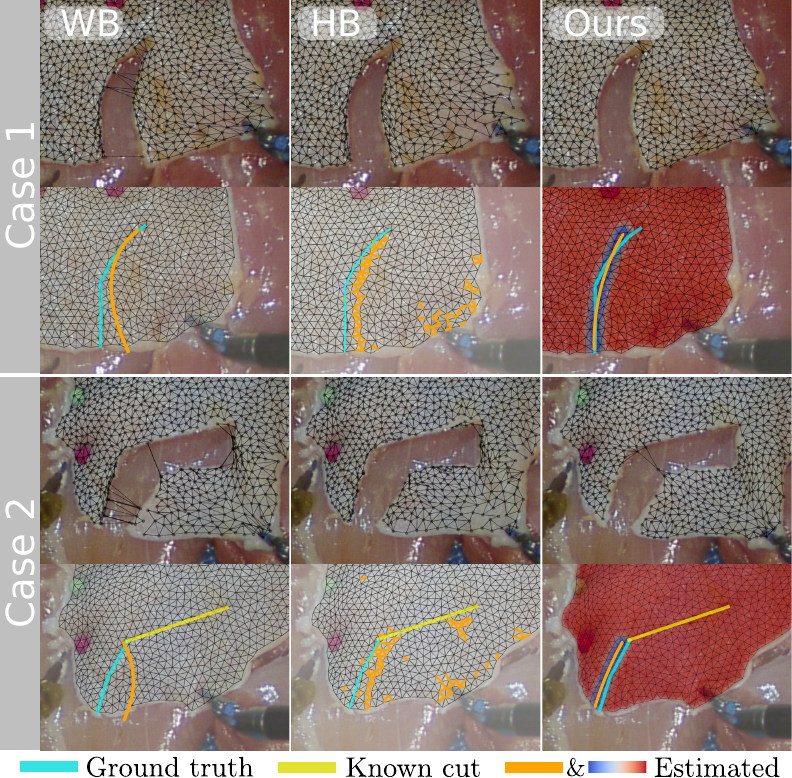}
    \vspace{-1.5em}
    \caption{Comparison of topology estimation results from three methods. \textit{WB} roughly predicts the location and shape but lacks accuracy due to wrong optical flow prediction. \textit{HB} is highly sensitive to observation noise, leading to many outliers. In contrast, our method achieves accurate topology estimation and enables the most realistic tissue state tracking.
    }
    \vspace{-0.5em}
    \label{fig:estimation_accuracy}
\end{figure}



\section{Experiments \& Results}

The performance of the proposed method is tested in an electrosurgery scenario where a real thinshell tissue is dissected by an cautery spatula within the da Vinci Research Kit (DVRK) \cite{dvrk}. We evaluate the proposed cutting curve estimation and feedback control algorithm seperately. Additional studies are done in simulated environment to analyze the proposed dissection pipeline.

\subsection{Baselines and Metrics}
The proposed method's estimation accuracy is compared against two adapted previous methods. They are
(1). \textit{WB}: a image warping-based baseline inspired by \cite{franccois2021image} that warps a hand-labeled incision mask to the reference frame with optical flow predicted from \cite{teed2020raft}. A quadratic \bbezier curve is found using least-square fitting to the warped mask. 
(2). \textit{HB}: a heuristic-based incision detection method similar to \cite{paulus2017handling}. It removes a mesh edge if it stretches more than a certain ratio threshold over the average of its neighbors during registration to point clouds.



Cutting topology estimation accuracy is evaluated using an Intersection over Union (IoU) metric
\begin{equation}
    \begin{split}
        \text{IoU}(\mathbf{R}_1, \mathbf{R}_2) = \frac{\mathbf{R}_1 \cap \mathbf{R}_2}{\mathbf{R}_1 \cup \mathbf{R}_2}
    \end{split}
\end{equation}
where $\mathbf{R}_1, \mathbf{R}_2$ are set of edges that are removed by estimation methods and manually labeled ground truth cutting curves. We also dilate the both edge sets to reflect the cases when two curves roughly matches but has small intersection.
For dissection trajectory accuracy, we compute the error as the Euclidean distance between corresponding positions sampled at uniform arc-length intervals along the desired trajectory and labeled resultant trajectory in canonical frame.

\begin{table}[t!]
\setlength\tabcolsep{1.5em}
\centering
\caption{Cut estimation accuracy of comparison methods in terms of Intersection-Over-Union (IoU) between dilated prediction and dilated ground truth cut.}
\begin{adjustbox}{width=0.48\textwidth}
    \begin{tabular}{l|ccc}
    \toprule
  & \textit{WB} & \textit{HB}  & ProCut (Ours) \\
  \midrule
Accuracy (IoU) &$0.54 \pm 0.23$ & $0.25 \pm 0.08$&$\mathbf{0.68 \pm 0.11}$ \\
\bottomrule
\end{tabular}
\end{adjustbox}
\label{tbl:estimation_accuracy_baseline}
\end{table}
\subsection{Estimation Performance}
Accuracy of comparison methods are estimating the dissect curve on a thin-shell tissue in an ex-vivo setting. In the experiments, a gripper tool retracts the pre-dissected tissue following a pre-defined trajectory to deform it. Note that, only in these experiments, the tissue is dissected before retraction so that ground truth cutting curves can be labeled. 

\begin{figure}[t]
    \centering
    \includegraphics[width=1\linewidth]{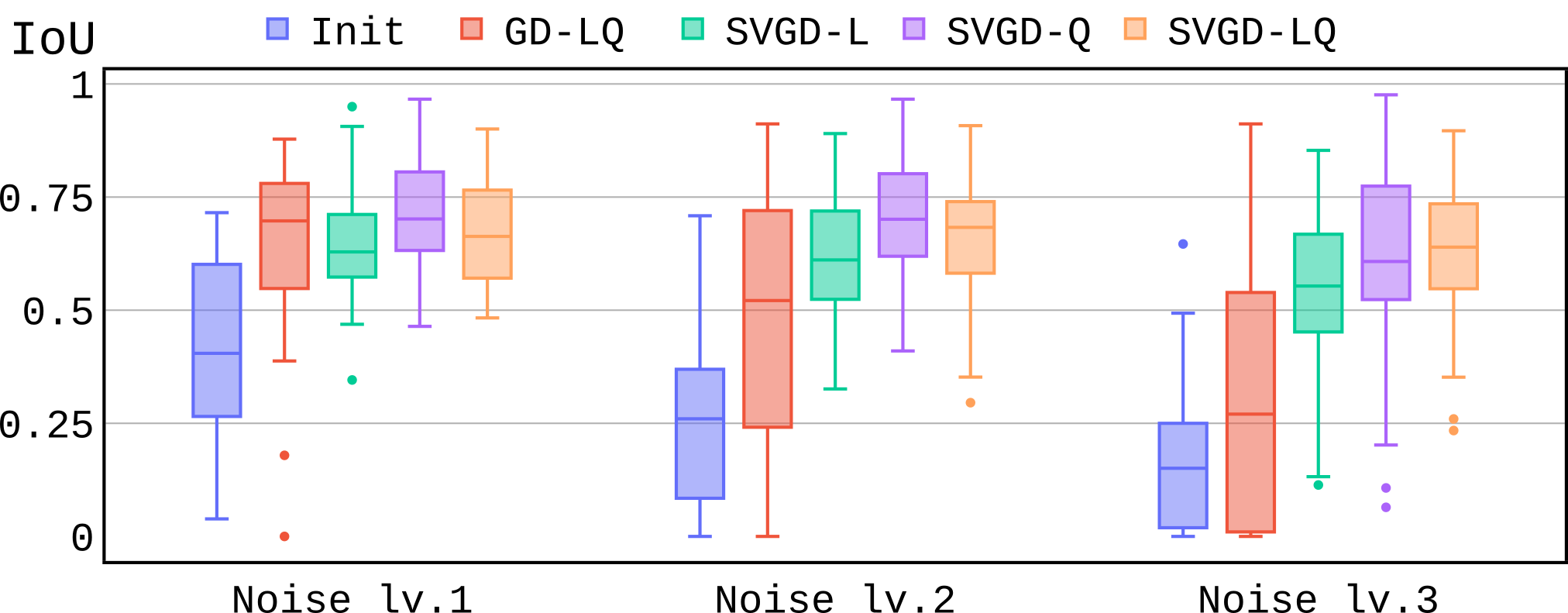}
    \vspace{-1.5em}
    \caption{Ablation results of comparing variants of our method initialized with different noise level. Data of 
\textit{Init} indicates initalization errors. Gradient-descent method (\textit{GD-LQ}) is sensitive to initialization. Estimation with linear representation (\textit{SVGD-L}) leads to lower accuracy, whereas quadratic representation alone (\textit{SVGD-Q}) is accurate but not robust to larger initialization error.  Combining SVGD with a linear to quadratic refinement strategy (\textit{SVGD-LQ}) achieves good accuracy and robustness.}
    \vspace{-1em}
    \label{fig:ablation_and_initialization}
\end{figure}

\autoref{tbl:estimation_accuracy_baseline} shows an overall accuracy of comparison methods on 5 different test cases involve real chicken skin tissue.
As our method relies on an initialization, we report its performance by averaging 10 experiments where uniform initialization noise of $\pm 5\ $mm is added by perturbing the start and end points of the ground-truth curve.
Overall, our method has on average the highest cutting topology estimation accuracy. This advantage is primarily attributed to the robustness of the gradient-based probabilistic estimation framework.
\autoref{fig:estimation_accuracy} illustrates representative results from the comparison methods.
\textit{WB} depends heavily on the accuracy of optical flow predictions and tends to fail when there are large pixel displacements.
\textit{HB} is capable of capturing both the position and shape of topological changes but is prone to generating outliers due to its sensitivity to observation noise.
In contrast, the proposed approach constrains the topology to a curve representation, enabling it to estimate topological changes robustly even in the presence of noise.

We evaluate the robustness of our method under varying initialization noise levels, with uniform noise of $\pm 5$ mm, $\pm 10$ mm, and $\pm15$ mm. As shown in \autoref{fig:ablation_and_initialization}, our method (\textit{SVGD-LQ}) maintains high IoUs of 0.68, 0.67, and 0.62, respectively. Ablation results confirm that combining SVGD with linear-to-quadratic refinement improves accuracy over linear-only (\textit{SVGD-L}), and SVGD offers greater robustness to large initialization errors compared to gradient descent (\textit{GD-LQ}) and using only a quadratic representation (\textit{SVGD-Q}).

\subsection{Autonomous Dissection Experiments}
We evaluate the proposed autonomous dissection algorithm in thin-shell tissue cutting task in both simulated and real-world scenarios. In both scenarios, a human operator command a desired cutting pattern either on the canonical space or on the image space. One tissue gripper retracts a thin-shell tissue in a pre-defined direction to tension it for ease of dissection, and a electrical cautery spatula executes the cutting trajectory. We assume that the robot produces error due to imperfect mechanical properties and inaccurate calibration \cite{liang2025differentiable}, which likely results in deviation of cutting trajectory without a proper feedback mechanism.

\begin{figure}[t]
    \centering
    \includegraphics[width=\linewidth]{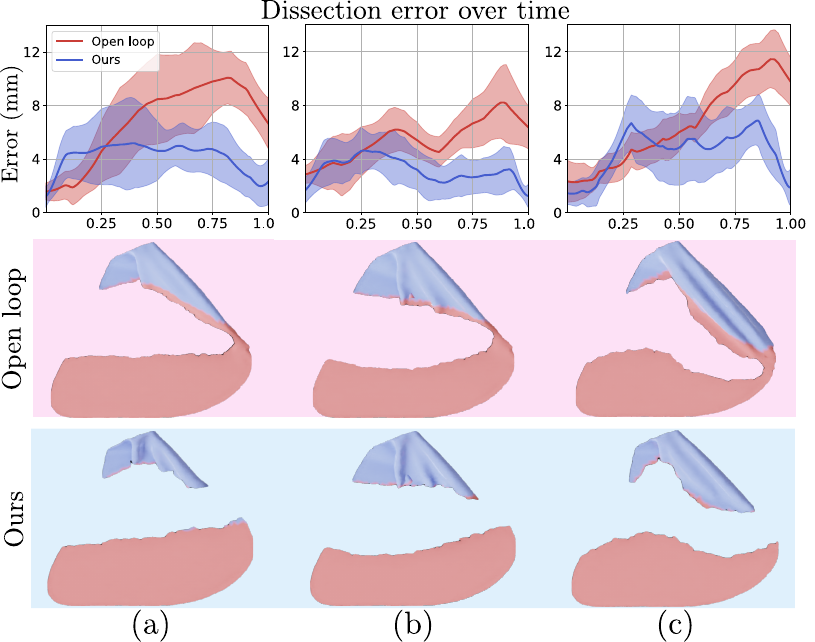}
    \vspace{-1.5em}
    \caption{Comparison of results from open-loop and the proposed method in simulation. (a)$\sim$(c) cases have target cutting patterns of \textit{line, arc, and snake} shape shown as the boundaries of two colors. The 1st row shows dissection error over time of 10 repetitions. The open loop method doesn't execute trajectory accurately and couldn't separate completely due to errors. The proposed method is able to follow desired pattern accurately.}
    \vspace{-0.5em}
    \label{fig:autonomous_cutting_sim}
\end{figure}
\begin{figure*}[t]
    \centering
    \includegraphics[width=1\linewidth]{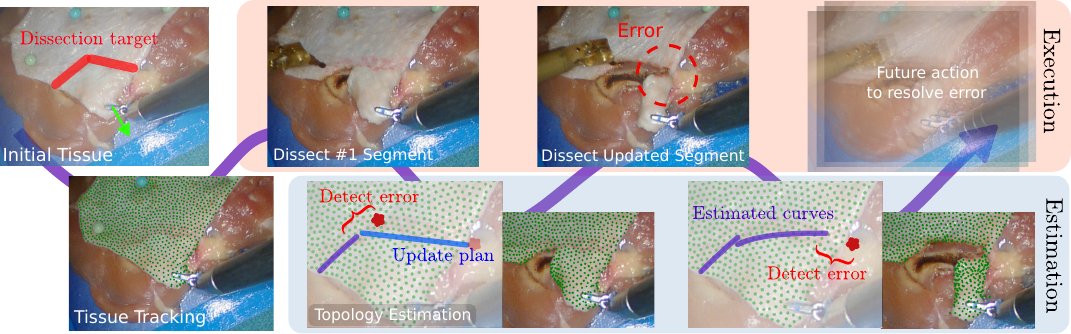} 
    \vspace{-1.5em}
    \caption{This figures shows an example flow of robot autonomous dissection using the proposed method. The robot utilizes the proposed method's tissue tracking capability to perform electrosurgical dissection. It inevitably makes error due to motion noise and calibration error. Our topology estimation method identifies the dissection error and updates the dissection plan. This process can be repeated until the dissection trajectory is met or when the estimated error is small enough. In this procedure, tissue retraction is tele-operated. 
    }
    \vspace{-1.5em}
    \label{fig:autonomous_dissection}
\end{figure*}
\textbf{Simulation Results:}
 We simulate robot kinematic error by a zero-mean Gaussian noise $p_e \sim \mathcal{N}(\mathbf{0}, \sigma_p^2\mathbf{I}_3)$ in robot end-effector position.
 A calibration error is introduced by a residual transformation $\mathbf{T}_e:\{\boldsymbol{\theta}_e, \mathbf{t}_e\}, \boldsymbol{\theta}_e \sim \mathcal{U}(-l_{\boldsymbol{\theta}}, l_{\boldsymbol{\theta}})^3,  \mathbf{t}_e \sim \mathcal{U}(-l_{\mathbf{t}}, l_{\mathbf{t}})^3$ to the robot base.
 Observation noise $z_e \sim \mathcal{N}(0, \sigma_z^2)$  is added to perturb every points in surface point cloud. We set $\sigma_{p} = 0.5 \text{ mm}, l_{\boldsymbol{\theta}} = \frac{\pi}{50}, l_{\mathbf{t}} = 2 \text{ mm}, \sigma_z = 0.5\text{ mm}$ in the simulated experiments. 
 We designed three dissection patterns with increasing level of complexities, they are \textit{Line, Arc, and Snake}. Results of comparing an open-loop method and the proposed close-loop dissection algorithm is shown in \autoref{fig:autonomous_cutting_sim}.
 The open-loop method is implemented from Alg.~\ref{alg:feedback_control} by removing its topology estimation and trajectory update.
 It deviates significantly from the desired trajectory and doesn't complete the dissection task due to noise and error. In comparison, our method is able to correct its dissection trajectory once it deviates from the desired one through estimation of cutting topology and tissue state, maintaining a lower dissection error over time. 
 Experiments for each dissection patterns are repeated for 10 times.

 \begin{table}[t!]
\setlength\tabcolsep{1.2em}
\centering
\caption{Results of real tissue dissection. \textit{Succ, Recov, Seg-Fail, Est-Fail} indicates successful dissection, estimation enabled recovery, failure due to image segmentation, and estimation failure.}
\begin{adjustbox}{width=0.43\textwidth}
    \begin{tabular}{c|cccc}
    \toprule
  \#\ Trials &Succ & Recov  & Seg-Fail & Est-Fail  \\
  \midrule
  10 & 50\% & 20\% & 20\% & 10\%\\
\bottomrule
\end{tabular}
\end{adjustbox}
\label{tbl:real_autonomous_cutting}
\end{table}

 \textbf{Real tissue Experiments}:
 In the real tissue dissection experiments, the robot autonomously dissect a piece of thin-shell chicken skin. 
 The autonomous dissection process is shown in \autoref{fig:autonomous_dissection}. 
 Each trial began with a human operator annotating the desired dissection path on an endoscopic image.
 Then they grasp and retract the skin tissue to create tension around the dissection area through tele-operation. 
During autonomous execution, our method continuously tracks the tissue’s pose to guide each cutting action, and after every dissection step it predicts the resulting cutting curve, detects any deviation from the plan, and updates the subsequent trajectory accordingly.
 The real-world experiments is repeated for 10 times and the results are shown in \autoref{tbl:real_autonomous_cutting} 
 A trial was deemed successful if the skin was cleanly separated into two pieces along the annotated path. 
 Our system achieved a 50\% success rate.
 Additionally in 20\% trials, the proposed estimation algorithm is able to detect that the tissue is not fully dissected, potentially enabling further dissection actions to fix the error. 
The remaining 30\% of trials failed outright, with 20\% attributable to large image‑segmentation errors and 10\% to inaccuracies in the cut‑curve estimation.
 In each experiment, a dissection target consists of two linear segments on a chicken skin tissue, with total length of around 5 centimeters.

\subsection{Implementation Details}
The proposed method is implemented in PyTorch and tested on a machine with RTX 3090 GPU. The tissue mesh contains 1000 and 2000 vertices in the real and simulated experiment respectively. All estimation experiments have 30 SVGD particles, and is ran for 200 steps. Dissection experiments have 10 SVGD particles and is ran for 100 steps. Linear representation is used for first half of the total steps and quadratic representation is used for refinement. An estimation step in a real dissection experiment takes around 0.3s.
The algorithm updates tissue state after every five topology estimation steps. Point cloud reconstruction of the tissue surface is achieved by using existing semantic segmentation \cite{SAM, CUTIE} and stereo disparity estimation \cite{RAFTSTEREO} techniques.

\subsection{Limitations}
While ProCut demonstrates robust closed-loop dissection on thin-shell tissue,
several limitations remain. First, our simulation models the mechanical
effects of cutting through localized stiffness relaxation but does not capture
the thermal and electrical physics of electrosurgery; consequently, the impact
of power settings, duty cycle, and carbonization on cut quality is not
predicted by the simulator and must be tuned empirically. Second, our real-tissue evaluation is limited to thin, homogeneous samples (chicken skin); extension to multilayer or heterogeneous tissues, such as fascia over fat or vessel-laden membranes,
would require modeling layer-wise stiffness and depth-aware cut termination,
which we leave to future work. Finally, tissue tensioning is currently
established once before cutting and held fixed; adaptive retensioning \cite{wang2025feedback} in
response to topology changes can be added to improve the dissection outcome.

\section{Conclusion} 
\label{sec:conclusion}
In this work, we presented ProCut, a novel framework that integrates differentiable physics and probabilistic inference to enable robust estimation of dissection topology in autonomous electrocautery procedures. 
By introducing a differentiable representation of cutting curves within a Position-based Dynamics (PBD) simulation and leveraging Stein Variational Gradient Descent (SVGD) for particle-based inference, our method effectively models and updates tissue topology.
Our method achieves superior accuracy in both simulation and real-world surgical settings, outperforming prior techniques in terms of robustness and precision.
We developed an autonomous dissection pipeline for thin-shell tissues that tightly couples cutting estimation with closed-loop feedback control. 
The system is capable of tracking and adapting to evolving tissue states, improving the reliability of robotic tissue dissection.

While our approach has been demonstrated on thin-shell tissues, the core idea of integrating differentiable physics with probabilistic inference for topology estimation is broadly applicable. By adapting the underlying topology representation, this framework can be extended to other scenarios, such as \cite{liang2024medic} and deformable attachments \cite{liang2024autopeel}. 
Additionally, the estimation algorithm can be enhanced to detect finer-grained topological changes and operate at higher update frequencies. This could be achieved through the use of dense visual tracking techniques 
 \cite{karaev2024cotracker} and more sophisticated GPU acceleration \cite{warp2022}.

 Our method lays the foundation for more advanced autonomous dissection and tissue manipulation tasks. Future work includes integrating our estimation framework with adaptive control policies and extending it to broader surgical scenarios. Ultimately, this approach has the potential to significantly enhance the precision and reliability of autonomous surgical procedures.



\bibliographystyle{ieeetran}
\bibliography{references}

\end{document}